\pdfoutput=1

\documentclass[11pt]{article}

\usepackage{EMNLP2022}

\usepackage{times}
\usepackage{latexsym}

\usepackage[T1]{fontenc}

\usepackage[utf8]{inputenc}

\usepackage{microtype}

\usepackage{inconsolata}
\usepackage{booktabs}
\usepackage{graphicx}
\usepackage{xcolor}
\usepackage{bbding}
\usepackage{pifont}
\usepackage{varwidth} 
\usepackage{array} 
\usepackage{xspace}

\definecolor{mygreen}{RGB}{43, 133, 64}
\definecolor{myred}{RGB}{205, 32, 38}

\newcommand{\ls}{Label Sleuth\xspace}
\newcommand{\lshomepage}{https://www.label-sleuth.org}
\newcommand{\workflowstep}[1]{\sectionitem{#1}}
\newcommand{\designstep}[1]{\emph{#1}}
\newcommand{\sectionitem}[1]{\textbf{#1}}

\newcommand{\vix}{\textcolor{mygreen}{\ding{51}}\textsuperscript{\textcolor{myred}{\kern-0.55em\footnotesize\ding{55}}}}
\newcommand{\vibw}{\ding{51}}
\newcommand{\ixbw}{\ding{55}}
\newcommand{\vi}{\textcolor{mygreen}{\ding{51}}}
\newcommand{\ix}{\textcolor{myred}{\ding{55}}}

\title{Label Sleuth: From Unlabeled Text to a Classifier in a Few Hours}

\usepackage[misc]{ifsym}

\author{Eyal Shnarch$^1$(\Letter)\thanks{\ \ These authors contributed equally to this work.}, Alon Halfon$^1$\footnotemark[1], Ariel Gera$^1$\footnotemark[1], Marina Danilevsky$^1$, Yannis Katsis$^1$, Leshem Choshen$^1$,\\
\bf{Martin Santillan Cooper$^1$, Dina Epelboim$^1$, Zheng Zhang$^2$, Dakuo Wang$^1$, Lucy Yip$^3$, Liat Ein-Dor$^1$,}\\ \bf{Lena Dankin$^1$, Ilya Shnayderman$^1$, Ranit Aharonov$^1$, Yunyao Li$^4$, Naftali Liberman$^1$,}\\ \bf{Philip Levin Slesarev$^1$, Gwilym Newton$^1$, Shila Ofek-Koifman$^1$, Noam Slonim$^1$, Yoav Katz$^1$} \\
$^1$IBM Research, $^2$University of Notre Dame, $^3$MIT-IBM Watson AI Lab, $^4$Apple\thanks{\quad Work done while author was working at IBM Research.}\\
\texttt{eyals@il.ibm.com}\\
}

\begin{document}
\maketitle
\begin{abstract}
Text classification can be useful in many real-world scenarios, saving a lot of time for end users. However, building a custom classifier typically requires coding skills and ML knowledge, which poses a significant barrier for many potential users. To lift this barrier, we introduce \textit{\ls}, a free open source system for labeling and creating text classifiers. This system is unique for (a) being a no-code system, making NLP accessible to non-experts, (b) guiding users through the entire labeling process until they obtain a custom classifier, making the process efficient -- from cold start to classifier in a few hours, and (c) being open for configuration and extension by developers.
By open sourcing \ls we hope to build a community of users and developers that will broaden the utilization of NLP models.
\end{abstract}

\section{Introduction} \label{sec:intro}

Text classification is an NLP task 
with great practical importance. 
Practitioners working with large amounts of textual data often need to 
categorize snippets of text.
For instance, a lawyer reviewing 
contracts may need to 
find clauses specifying the terms under which a contract can be terminated.
Or, a psychologist analyzing treatment notes may be interested in finding all sentences that indicate that a patient is suffering from depression. 
Often, 
the text snippets of interest are rare and scattered throughout the 
dataset. Manually reviewing the entire dataset is inefficient or impractical, thus raising the need for an automated solution in the form of a custom text classification model.

Practitioners, or \textit{domain experts} (a.k.a subject matter experts)
who need such models, typically 
lack the 
skills to build them, and thus must rely 
on Machine Learning (ML) experts. 
This, in turn, creates a gap between modern text classification techniques and their end users, which we aim to bridge in this work.

We present \href{\lshomepage}{\ls}\footnote{\url{\lshomepage}} -- an \href{https://github.com/label-sleuth/label-sleuth}{open source}\footnote{\url{https://github.com/label-sleuth/label-sleuth}} system designed to enable domain experts to create a text classifier by themselves, 
with no dependency on ML experts. \ls is both a labeling platform and a machine learning platform,
and is thus 
used to collect labeled data as well as to build text classifiers. 
It enables a domain expert to create a 
good quality custom classifier from a cold start (no labels)  in a few hours, in several short rounds of labeling -- enhanced via active learning \citep{cohn1996active} -- that provide feedback to models being trained in the background. 
\ls was designed to be intuitive and easy to use by domain experts. Rather than trying to cover many different NLP tasks and increasing the system complexity, it focuses on a single broadly applicable use case of binary text classification, and provides a fully automated flow for building such classifiers.

To the best of our knowledge, \ls is the first text classification platform intentionally 
designed for a broad audience - domain experts that typically lack coding skills or an understanding of ML concepts. By open sourcing \ls, we hope the community will join this effort, to further expand and improve its existing capabilities for the benefit of a potentially wide community of users. 
\section{System description} \label{sec:sys}

\begin{figure*}[th]
\begin{center}
\includegraphics[width=1\textwidth]{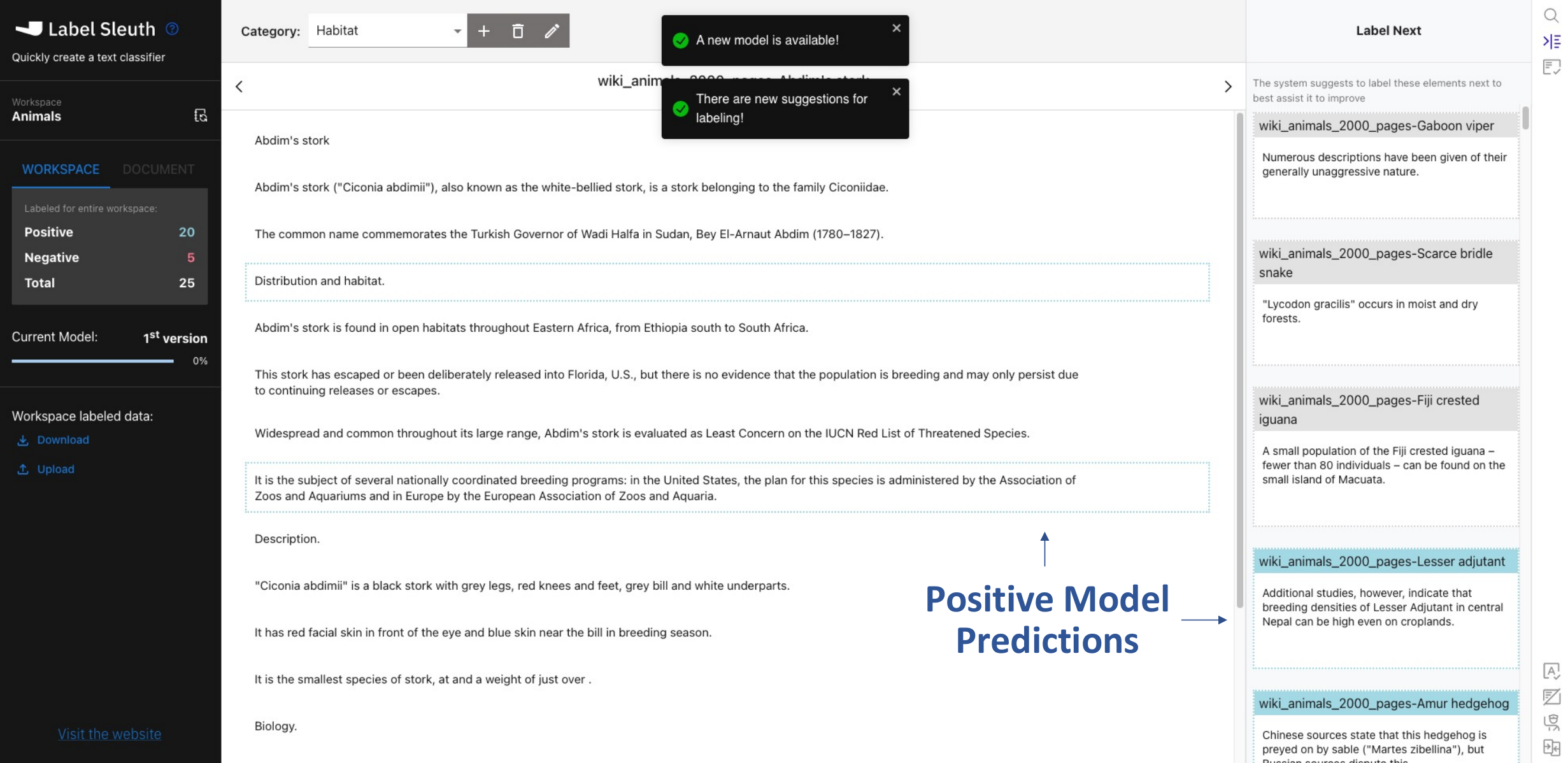}
\caption{\textbf{The workspace screen} when the first classifier is ready. \ls notifies the domain expert that a new model is ready and guides her to the \textit{Label Next} list in the right panel. The sentences that are predicted as positive by the new classifier are marked with a blue frame (both in the document view and in the list on the right).  
}
\label{fig:next_to_label}
\end{center}
\end{figure*}
 
\subsection{A typical workflow} \label{ssec:workflow}

We illustrate a flow for using \ls through the eyes of a potential user. We encourage readers to experience this workflow directly, to get a first-hand impression of the process.\footnote{A step-by-step \href {https://www.label-sleuth.org/docs/tutorial.html}{tutorial} is provided on the website.}
Consider Viki, a Wikipedia editor and expert in animals, who is interested in enriching the content of animal articles on Wikipedia.
Her goal is to ensure that accurate information about an animal's habitat is included in all animal articles. 
Manually reviewing all articles would be extremely grueling.
Instead, she can use \ls to build a custom binary classifier for this task. The classifier will identify sentences describing an animal's habitat, allowing her to focus on relevant sentences to review, and to identify articles missing habitat information.

\workflowstep{Upload data and create a classification category.} 
To get started, she uploads the set of Wikipedia articles, split into sentences, into the system.\footnote{The \ls installation includes this dataset.} 
She then creates a new \textit{workspace} using the uploaded 
corpus. Figure~\ref{fig:next_to_label} depicts \ls's workspace screen: in the center is a document view; the left panel presents information about the status of the labels and model; and the right panel is populated with various lists of  text examples from the corpus (more details below). 
The workspace enables her to 
create multiple custom categories (i.e., classes) for classification. Based on her needs, Viki creates a \textit{Habitat} category and starts labeling sentences as belonging (or not belonging)
to it.

\workflowstep{Finding examples to label.}
Viki can use the document view to skim 
articles and label sentences. However, since 
 sentences about habitats are relatively rare, this would lead to her mostly labeling negative examples.
To quickly find positive examples, Viki leverages \ls{}'s search functionality.
Based on her domain knowledge, she thinks up some relevant terms -- for instance, the category name \textit{habitat} or the phrase \textit{lives in} -- and uses the \emph{Search} option on the right panel to retrieve a list of sentences 
that mention these terms and thus are 
more likely to belong to the category.
Search results can be labeled directly using the \vibw\ and \ixbw\ buttons. 
If an example's context is needed to make a decision, clicking on it shows the source article in the document view, highlighting the example.

If Viki has already collected some labeled examples outside \ls{}, she can bring them into the system with the \textit{Upload} option on the left panel.

\workflowstep{Automated model training.}
Once a sufficient set of labeled examples 
is provided (see App.~\ref{app:default_config}), \ls{} automatically starts 
training a classifier in the background. Viki does not need to manually invoke training. 
However, she can use the progress bar on the left panel to track her progress and see how many more labels are needed before the system starts training a new classifier.

\workflowstep{Receive guidance on what to label.} Once the first classifier is ready, 
\ls 
leverages it to identify unlabeled examples that would be most beneficial to label next, using 
an active learning strategy \citep{cohn1996active}. It then populates a new \emph{Label Next} list with the selected 
examples in the right panel, and invites Viki to label this list. 
Fig. \ref{fig:next_to_label} depicts the system when the first classifier is available.
As Viki keeps labeling, \ls triggers a new iteration, in which a new classifier is trained, and its predictions and the Label Next list are updated accordingly. With the additional labeled examples, the classifier improves with each such iteration.

\workflowstep{Review model predictions.}
At any point, Viki can review the predictions of the current classifier to get an impression of its performance 
and provide feedback. She can do this by skimming through different articles in the main document view, which have the positive predictions highlighted. Alternatively, she can open the \textit{Positive Predictions} list on the right panel to see the sentences, across all articles, that received a positive prediction. If she disagrees with a prediction, she can directly label the corresponding element to provide focused feedback 
to the model. 

\workflowstep{Evaluate model quality.} To get a more concrete measure of the classification quality, Viki can initiate  
a \textit{Precision Evaluation} procedure. The system samples $n$ 
sentences that are predicted as positive by the current classifier. Viki 
is asked to label these sentences and 
her feedback is used to estimate the precision of the classifier.

\workflowstep{Receive guidance on potential labeling errors.}
While working on a repetitive labeling task, it is natural to make mistakes. 
These mistakes introduce noise to the labeled data,
resulting in degraded model performance.
To mitigate this, \ls identifies and surfaces potential labeling errors for Viki to review and correct as needed (see Appendix~\ref{app:labeling_reports} for details). Identifying labeling errors and understanding their causes early on not only improves the performance of the resultant classifier but can 
also sharpen the user's understanding of the task for future rounds of labeling. 

Finally, once Viki is satisfied with the classifier performance, she 
can continue her review inside \ls, rapidly reviewing the articles she has uploaded (or new articles that she can upload at any time), focusing on the sentences predicted by the classifier, and making sure that habitat information is present and correct.\footnote{Users with engineering skills may export the classifier created by the system and use it on their own environment, or download the collected labeled data and use it to train a different classifier.} 

\subsection{Guiding design principles}

\ls{} is designed to enable domain experts to build custom text classification models. This is in stark contrast to alternative systems that focus on technical users, be it data scientists or ML experts (see \S~\ref{sec:related}). 
We next describe the main principles guiding the design of \ls, in the context of the above workflow.

\sectionitem{Minimize the labeling effort.} 
The time of domain experts is typically limited and expensive. 
The system should thus make effective use of their time, as well as demonstrate a quick return on investment to keep them engaged. 
\ls accomplishes this in the following ways:

\designstep{Focus on value-added positive examples.} When it comes to building a text classifier, not all labels are equally important. For instance, in the common case where positive examples are scarce, it is these positive examples that are more valuable. Therefore, \ls{} initially guides domain experts towards identifying a seed of positive examples through its search functionality. Since negative examples are more common, \ls{} does not force the user to label them. If the domain expert has not provided enough negative examples to train a model, \ls{} automatically completes the missing info by randomly selecting unlabeled examples to be considered as weak negative examples, thus reducing the domain experts' labeling effort.

\designstep{Continuous labeling guidance.} As the flow progresses, \ls{} further ensures that domain experts focus on labeling important elements by continuously guiding them through the labeling process. 
This guidance comes in two forms. First, by providing active learning suggestions, the system focuses domain experts on labeling examples useful for improving the model, instead of wasting effort on labeling uninformative examples. Second, by providing label error analysis, \ls{} allows users to promptly catch issues with their labeling (e.g., caused by concept drift or ill-defined categories) and revise their work before wasting more time on erroneous labeling.

\designstep{Progress feedback.} Finally, to further reduce user effort, \ls{} provides continuous feedback on the model performance. By showing the classifier's predictions, as well as via the precision evaluation mechanism, the system enables users to understand when the classifier's performance is adequate and they can safely stop labeling.

\sectionitem{Abstract the ML process.} Domain experts, while proficient in their domain, may not be familiar with ML techniques or even ML terminology. As a result, the system should abstract the ML process as much as possible. This is accomplished in \ls{} through the following features:

\designstep{Automated and transparent ML processes.} 
All ML steps, including model training, inference, and active learning, are automatically initiated
and performed in the background without user intervention. Once completed, a colorful confetti animation notifies the user that a new classifier is ready (this also serves as a surprisingly effective means for keeping users engaged). Other than being aware of the classifier being iteratively trained by the system, users are not expected to have any ML knowledge.

\designstep{Out-of-the-box configuration.}
\ls users do not have to worry about setting various parameters, e.g., choosing the model architecture or active learning strategy. The default system configuration defines a workflow that suits a typical classification use case (see App.~\ref{app:default_config}). While more advanced users can easily change and adapt the configuration (see \S \ref{sec:architecture}), the emphasis is on having a hassle-free setting that is available out-of-the-box.

\subsection{Real usage examples}
Several early users have already successfully applied \ls to their real-world tasks.
For instance, a legal user needed a text classifier to identify clauses of interest in long contracts. After working for 6 hours on \ls, they built a classifier for a category of high-risk clauses. 
By highlighting relevant clauses for review, where they would otherwise have needed to review contracts in their entirety, they estimate \ls to have saved them 80\% of their time.

In another example, \citet{gretz2022benchmark} developed VIRA, a chatbot that helps address COVID-19 vaccine hesitancy. They relied on \ls to build a dialogue act classifier, which maps user chat utterances into general categories (e.g., \textit{greeting, query, concern}); these are used to determine whether to reply to the user with a corresponding generic response, or to pass the utterance to a dedicated intent classification system. 
VIRA researchers testify that besides the label collection itself, \ls was valuable in helping them fine-tune the definitions of target categories and converge on their desired classification task.

The latter example shows how \ls{} can be useful for ML experts; it provides a method to quickly obtain auxiliary classifiers needed for intermediate steps, and enables ML experts focus their attention and time on the larger tasks.

\section{System comparison} \label{sec:related}
Text labeling (or annotation) tools have proliferated in recent years. \citet{neves2021extensive} surveyed 78 tools. They can be classified into two categories:

\textbf{Basic labeling tools} simply allow users to assign a label(s) to data elements.
Examples include early tools, such as Callisto \cite{day2004callisto}, BRAT \cite{stenetorp-etal-2012-brat}, and WebAnno \cite{yimam-etal-2013-webanno}, and more recent ones, such as Doccano \citep{doccano}.

\textbf{Labeling tools with ML support} are more similar to \ls, since in addition to collecting labels, they train a classifier with these labels, or accelerate the labeling process by integrating ML.

In our comparison, we focus on representative systems that are most similar to \ls{} and have gained users popularity.
All these systems offer some form of ML labeling support, However, they are designed with technical users in mind, such as data scientists and developers; they often require complex actions to get started, which assume ML knowledge. 
They do not offer ML integration out-of-the-box, relying instead on the user to configure the system (e.g., by connecting it to external models). Furthermore, ML support is typically limited to active learning and lacks
advanced features that could help domain experts, such as identifying and guiding the user in resolving potential labeling issues. 
We next provide a brief overview of each of the reviewed systems. Table \ref{tab:competitors} summarizes their features compared to \ls.

\begin{table*}
\small
\centering
\resizebox{\textwidth}{!}{%
\begin{tabular}{@{}lcccccc@{}}
\toprule
 & \begin{tabular}[c]{@{}c@{}}No technical \\  expertise needed \end{tabular} &
\begin{tabular}[c]{@{}c@{}}ML guidance on \\ what to label \end{tabular} &
\begin{tabular}[c]{@{}c@{}}ML guidance on \\ label errors \end{tabular} &
 \begin{tabular}[c]{@{}c@{}}Open \\ source\end{tabular} & 
 \begin{tabular}[c]{@{}c@{}}Tasks other than\\ text classification\end{tabular} &\\ 
\midrule
Prodigy & \ix & \vi & \ix & \ix & \vi\\
Label Studio (Free)  & \ix & \ix & \ix & \vi & \vi\\
Label Studio (Paid)  & \ix & \vix & \ix & \ix & \vi\\
INCEpTION & \ix & \vix &  \ix &  \vi  & \vi \\
\midrule
\ls{} & \vi & \vi & \vi & \vi & \ix &\\
\bottomrule
\end{tabular}%
}
\caption{\textbf{Comparing \ls} with representative text labeling tools with ML support. The \vix~sign denotes cases where the functionality exists but is very complicated to set up. 
}\label{tab:competitors}
\end{table*}

\emph{Prodigy} \citep{montani2018prodigy} is a paid, closed source labeling tool by the makers of \href{https://spacy.io}{spacy}. While it offers an intuitive frontend, it targets mainly data scientists, as most tasks (except for basic labeling) - including dataset upload - require using the command-line. Moreover, it does not show examples in context and thus the user must label them in isolation from their source document, and according to a predefined order.

\emph{Label Studio} \citep{Label_Studio} is offered in a free open source community edition and a paid enterprise edition. While the latter offers ML and active learning support, setting up the process requires invoking external models (which in their simplest form are pre-built container images).

\emph{INCEpTION} \citep{tubiblio106270} -- an open source labeling tool from TU Darmstadt --  is arguably the most configurable tool in the list. It enables fine-grained control of several aspects of the labeling process, including the label granularity and when model predictions are shown.
However, this customizability further increases the barrier to entry compared to other tools. Even setting up a classification task requires creating complex annotation layers, while integrating a model, in many cases, requires the use of external libraries.

\medskip
In contrast to \ls{}, these systems support NLP tasks other than text classification, such as NER and question answering, or even non-textual tasks, such as audio and image classification. Thus, \ls{} and these systems correspond to different points on the trade-off between ease of use and task support. Existing systems support a wide variety of tasks but assume a technical user, while \ls{} focuses on text classification but creates an end-to-end model building experience tailored specifically to non-technical users. We believe that it is important to have tools that strike different balances in this trade-off.

\section{Architecture} \label{sec:architecture}
\ls is composed of backend and frontend layers. The backend is written in Python and uses the Flask framework for exposing a web service; the frontend is a React application which uses the MUI design library. For additional details see our \href{https://www.label-sleuth.org/docs/dev/architecture.html}{architecture webpage}.

While \ls 
is well-suited to users with no ML background, it also offers configurability and extensibility options for advanced users.
Users can choose from the available \textit{models} and \textit{active learning strategies}, and can also contribute new ones by implementing one or two straightforward functions. In addition, it is possible to configure the system to dynamically switch between models and/or strategies as the labeling progresses. 
Large models that require a GPU are also supported.

Another extensible component is \textit{training set selection}. While a basic approach would be to train classifiers using the set of examples labeled by the user, more advanced methods can provide added benefits. The default setting leverages the fact that the negative prior is high (since positive examples are relatively rare), and randomly selects elements from the unlabeled set to be added as weak negative examples for training.

The various system configurations (e.g., classification model, active learning strategy, criterion to trigger the training of a new model) together constitute a policy that shapes the flow and experience of building a classifier. The default policy (see App.~\ref{app:default_config}) can be extended and modified to further improve efficiency or to support different scenarios.

The \textit{data access} layer is responsible for saving and exposing the dataset and user labels. The current implementation relies on a combination of in-memory for performance, and local disk storage for persistency.
Finally, while English is used as the default language, \ls provides an infrastructure to easily support other languages.

\section{Open source and research opportunities} \label{sec:os}

\ls is the product of a collaboration between industry and academia, and aims to continue evolving by leveraging insights from multiple stakeholders and perspectives. We welcome further contributions and feedback from domain experts and the open source community, as well as researchers in related fields, including natural language processing and human-computer interaction.

To facilitate this, \ls{} was released in July 2022  as open source under the Apache 2.0 license. Following the example of other successful projects, in addition to the  \href{https://github.com/label-sleuth/label-sleuth}{source code} of the system, the open source release includes  material aimed to facilitate the use of the system and contributions to its development. The material on the \href{\lshomepage}{project's website} includes an overview of the system (including a short video), quick installation instructions, and a walk-through tutorial tailored to domain experts (building upon the animal habitat scenario and dataset of \S~\ref{sec:sys}). There is also detailed documentation of the system's internals for open source contributors and/or researchers who want to understand the underlying techniques and extend the system for their own needs.

As detailed in \S~\ref{sec:architecture}, the system is highly extensible, allowing researchers to further improve the system by incorporating novel techniques. A research aspect that we believe will be of particular interest to the NLP community is the unique requirements that arise from the interactive nature with the non-technical target audience of \ls{}. We next outline a few examples of such requirements, which we hope the NLP community will contribute solutions to.

\begin{figure}[t]
    \centering
    \includegraphics[width=0.8\columnwidth]{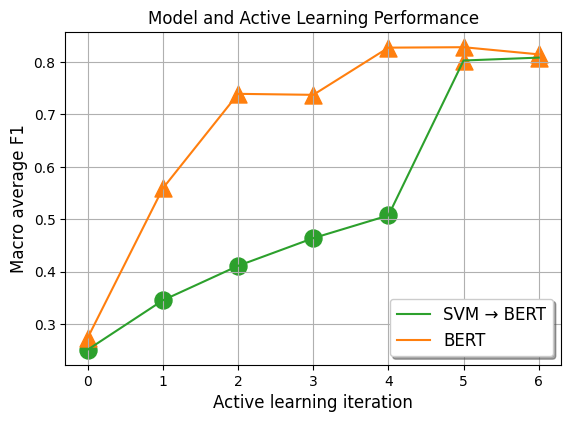}
    \setlength{\belowcaptionskip}{-10pt}
         \caption{
         \textbf{Policy setting: Choosing the classification model.} If the model in the last two iterations is BERT, using the lighter SVM for the first four iterations does not harm performance in comparison to using the heavier BERT for all iterations.
         Each point represents the avg. F$_1$ over 5 classes from 5 different datasets and 5 repetitions (seeds). Each  iteration adds 30 examples. See App. \ref{app:experiment} for details.}
         \label{fig:svm-vs-bert}
\end{figure}

\sectionitem{Setting the policy.}
One such challenge is the selection of employed policy (ML models, active learning techniques, etc.). For instance, consider the choice of classification model. In a static system with no interaction, performance on the task may be the most important model characteristic, and thus a large (and slow) model may be preferred. However, in an interactive system like \ls{}, lightweight and fast models have some unique advantages, providing faster turnaround time and thus more immediate feedback and guidance. Initial experiments, depicted in Fig.~\ref{fig:svm-vs-bert}, show that utilizing a light SVM model for most iterations and only switching to the heavier and high-performing BERT model \cite{BERT} for a few final iterations, leads to an F$_1$ score that is comparable to using BERT the entire time, while offering a significantly faster run time which improves the interactivity experience.

\sectionitem{Model evaluation.} Another example
is model evaluation. In typical NLP experiments, performance is quantified using some metric (e.g., F$_1$) over a test set. However, this differs from the needs of a typical \ls user in two ways.

First, maintaining a separate test set, which is not utilized for model training, undermines the goal of minimizing the labeling effort. 
Cross validation evaluation is incorrect in this scenario as the labeled examples collected in the process are not necessarily a good representation of the data (being biased towards positive examples and by active learning suggestions).
As an initial solution, after the model performance is estimated via the \textit{Precision Evaluation} process (\S\ref{ssec:workflow}), the examples that had been labeled for this purpose are added to the training set. In addition, estimating metrics such as recall is impractical when the positive prior is low (common in real-world classification tasks), since a reliable estimate requires a very large amount of test labels.

Second, a very important aspect is communication of evaluation results, especially in such an interactive system. Domain experts want to understand the performance of their classifier, but quantitative metrics such as F$_1$ may not be intuitive to them \cite{kay-chi-2015}. Thus, there are research challenges for both finding metrics that are less data-hungry, and constructing a user experience to best reflect model performance and convey a tangible sense of progress.

\sectionitem{Warm start.} Last but not least, advances in pretrained language models and in zero-shot text classification (e.g., \citealp{yin-etal-2019-benchmarking}) open up new opportunities to jump-start the process of building a classifier. However, integrating such techniques into \ls{} requires understanding the inputs expected by these techniques (e.g., category names or descriptions) and how to acquire them from domain experts. Moreover, work is needed to combine zero-shot with supervised techniques into a natural user workflow, where users not only get a good initial model (through zero-shot techniques), but also have the ability to further improve it by providing additional feedback.

\section{Conclusions} \label{sec:conclusions}
\ls is a production-ready freely available open-source system that seeks to lower the accessibility barrier for domain experts to label and build text classifiers.
It provides unique opportunities for a more productive and efficient classifier building process -- one where the system guides the labeling process, and both the domain expert and the ML components can provide timely feedback to each other.
We encourage domain experts, the open source community, and researchers to use, extend, and contribute back to the \ls project.

\section*{Limitations}
As mentioned in \S~\ref{sec:related}, being focused on a single task has its limitations. The obvious one is not supporting other useful tasks, such as entity labeling, relation extraction, question answering. Building a version of \ls dedicated for another task will demand the effort of redesigning the workflow and the interaction with the users.

In the text classification realm, \ls is limited in the type of task it handles -- a binary classification. Thus, in the case of a multi-variate category, such as \textit{Emotions} which may include several labels (e.g., joy, fear, anger, sadness), working with \ls demands creating a binary category for each of the labels. 

In the case of mutually exclusive categories, one can export all labeled data and train a multi-class classifier outside \ls. However, if the categories are not mutually exclusive, the selected data cannot be used as is for training a multi-label classifier, as it is likely that most collected examples were only labeled for a subset of the categories. 

One way \ls reduces the labeling effort is by minimizing the number of negative examples needed. The system achieves this by automatically selecting unlabeled examples as weak labeled examples, relying on the low prior of positives. If this is not the case, this feature should be disabled and users would have to spend additional time on labeling negative examples.

Finally, \ls requires that the uploaded documents are split into text elements. 
This split is static once the data was loaded. Thus, users are limited to labeling these standalone elements. They cannot, for example, mark that several elements constitute a positive example only when considered together. This requirement stems from the need to perform inference during the labeling process, which in turn requires specifying the text units to be inferred.

\section*{Ethics Statement}

We believe that this work has the potential to make NLP model building more inclusive by making it accessible to community members that until now did not have the means to create custom models; whether that was due to lack of technical knowledge or due to lack of resources to hire ML experts. 
At the same time, there are important ethical issues that should be considered and taken into account in the design, implementation, and use of \ls.

First, since the goal of the system is to automate parts of the model building process, it has the potential to take over responsibilities that were until now carried out mainly by ML experts/developers. While this is an important issue whose effects should be carefully considered and mitigated, we should note that ML experts could be involved in the process in new ways, such as: (a) by participating in the design, implementation, and extensions of the system itself, and (b) by leveraging the labeled data collected by \ls{} to build even more sophisticated ML models.

Second, since \ls{} is designed and implemented by humans and interacts with humans, there is potential for the introduction of bias. Bias could be introduced in two main places:

\emph{System design and implementation:} Design and implementation decisions made by developers of the system may introduce unwanted bias. This includes decisions on the frontend (e.g., using culture-specific icons, supporting only left-to-right languages on the frontend, etc) and the backend (e.g., selecting model learning algorithms that support or perform better in specific languages, etc.). We will be working with the \ls{} contributors to restrict such design bias as much as possible.  

\emph{Data, annotations, and model:} Bias can also be introduced into the learned model as a result of information provided by the domain expert, including the uploaded text data and provided labels. To avoid such bias, the system should inform the domain expert of potential implicit bias and suggest ways to mitigate it (such as uploading more diverse datasets). Understanding how to identify, communicate, and allow domain experts to limit such bias is a very interesting area of future research.

Finally, \ls{} inherits all considerations that apply to the use of ML models, including understanding their limitations and avoiding blind trust. This is partially mitigated by the fact that \ls affords the user an opportunity to discover and fix model issues quickly within the system, as opposed to other ML applications where the model is static and the user has a limited ability to affect the model.

\section*{Acknowledgements}
We thank Frederick Reiss for his guidance on open-sourcing the project, Natti Eder and Shao Zhang (Shanghai Jiao Tong University) for their contributions to the system design, and Lucian Popa for supporting the project.

\bibliography{LSdemo}
\bibliographystyle{acl_natbib}

\appendix

\begin{table*}
\centering
\begin{tabular}{@{}cccc@{}}
Dataset & Target category & Query & Test size \\ 
\toprule
20 Newsgroup & sci.med & 'health | medicine' & 7532 \\
AG News & World News & OR over a list of countries and territories & 3000 \\
DBPedia & Company & 'company' & 3000 \\
ISEAR & Joy & 'joy | happy' & 1534 \\
Yahoo! Answers & Sports & 'sports' & 3000 \\
\end{tabular}%
\caption{Dataset used in the experiment whose results are presented in Figure \ref{fig:svm-vs-bert}.}
\label{tab:datasets}
\end{table*}

\section{Default policy} \label{app:default_config}

As explained in \S~\ref{sec:architecture}, the various configurations of \ls form a \emph{policy}, which controls the flow of the model building experience.  \ls{} offers a default policy suitable for domain experts (which can be further modified by advanced users as needed).
Below, we list the default policy of \ls, as of its initial open source release. While we do not claim that the chosen settings are optimal, they were chosen empirically, by conducting multiple experiments on a wide variety of use cases and have been found to work well for typical text classification use cases.

\emph{Training invocation:} \ls{} starts training the first classifier once $20$ positively labeled examples are collected. After the first classifier, a new classifier is automatically trained for every $20$ new labels (positive or negative) provided by the user.

\emph{Training set selection:} Leveraging the low-prior scenario, the system can add unlabeled elements as weak negative examples for training. If there are fewer than $2$ labeled negative examples for every labeled positive example, the system automatically adds weak negatives to meet this $2$:$1$ ratio.

\emph{Precision evaluation:}
Whenever the user invoke a precision evaluation procedure, the system sample $50$ examples which are predicted as positive by the current model and asks the user to label them. Once labeled, the system can report precision and add these newly labeled examples to the training set to be used by subsequent training iterations. 

\emph{Machine learning algorithm:} The default classifier is an ensemble of two SVM \citep{svm95} classifiers -- one using Bag-of-Words representations and the other using GloVe \citep{pennington2014glove} representations.

\emph{Active learning strategy:} The default active learning strategy is uncertainty sampling \citep{lewis1994sequential}.

\section{Labeling quality analysis} \label{app:labeling_reports}
\ls currently employs two approaches to surface potential errors and inconsistencies in the labels provided by the domain expert. Each approach yields a list of labeled elements, which is then presented to the domain expert to review and correct as needed.

In the first method, the list of elements to review is based on disagreements between classifier predictions and user labels. The classifier was given these labels as training examples, which presumably lowers the chance of such direct disagreements. Therefore, the implementation relies on cross-validation: several classifiers are trained on different parts of the labeled data; if a classifier's prediction on a left-out element disagrees with the user-provided label for that element, it is added to the list for review. This list is sorted according to the classifier's confidence score.

In the second approach, the system presents pairs of examples that have been assigned contradicting labels w.r.t. the target category by the domain expert even though they are semantically similar to each other. This raises the possibility that one element in the pair was given an incorrect label. The list of pairs to be reviewed by the user is sorted based on decreasing similarity. In the current implementation, similarity is calculated by the distance between the average GloVe \cite{pennington2014glove} embeddings of the two texts.

\section{Figure \ref{fig:svm-vs-bert} experimental details} \label{app:experiment}

Below we describe the setting for the experimental results shown in Figure \ref{fig:svm-vs-bert} and described in \S~\ref{sec:os}.

We experiment with the use of different models over 6 active learning iterations. In each iteration, training examples are added using uncertainty active learning \cite{lewis1994sequential} over the previous model predictions. We compare two settings: One setting uses  a BERT classifier for all iterations, while the other uses SVM for iterations 0-4 and BERT for iterations 5-6 only.

Iteration 0 starts with a query tailored for the target class. Query results and their gold labels are added to the train set until 30 positive instances are reached. These query instances are used to train the iteration 0 model. In each subsequent iteration, a batch of 30 examples, selected by the active learning strategy, is added to the train set and a new model is trained.

Experiments were performed on one target class from each of the following 5 datasets: 20 Newsgroup \cite{ds-20-newsgroup}, AG News \cite{ds-ag-news-dbpedia-yahoo}, DBPedia (\citealp{ds-ag-news-dbpedia-yahoo}, CC-BY-SA), ISEAR (\citealp{ds-isear:15}, CC BY-NC-SA 3.0) and Yahoo! Answers \cite{ds-ag-news-dbpedia-yahoo}. Each experiment was repeated 5 times, using different random seeds for sampling from the query. Class and query details appear in Table \ref{tab:datasets}.

The active learning experiments were run with the \href{https://github.com/IBM/low-resource-text-classification-framework}{Low-Resource Text Classification Framework} repository \cite{BERT-AL-EMNLP20}, using their train-dev-test splits. For BERT, we fine-tuned BERT\textsubscript{BASE} (110M paramaters) for $5$ epochs, with a learning rate of $5 \times 10^{-5}$ and batch size 32. For SVM, we used the \href{https://scikit-learn.org/}{scikit-learn} Linear SVC implementation over Bag-of-Words representations (using CountVectorizer with max\_features=10000). In total, the experiment included 175 BERT fine-tuning and inference runs, equaling about 12 total GPU hours using a Tesla V100-PCIE-16GB GPU.

\end{document}